\documentclass{article}

\usepackage[final]{nips_2018}

\usepackage[utf8]{inputenc} % allow utf-8 input
\usepackage[T1]{fontenc}    % use 8-bit T1 fonts
\usepackage{hyperref}       % hyperlinks
\usepackage{url}            % simple URL typesetting
\usepackage{booktabs}       % professional-quality tables
\usepackage{amsfonts}       % blackboard math symbols
\usepackage{nicefrac}       % compact symbols for 1/2, etc.
\usepackage{microtype}      % microtypography
\usepackage{float}
\usepackage{graphicx}
\usepackage[titletoc,title]{appendix}

\title{Improving Robustness of Neural Dialog Systems in a Data-Efficient Way with Turn Dropout}

\author{
  Igor Shalyminov\thanks{The work was done during an internship at Microsoft Research}\\
  School of Mathematical and Computer Sciences\\
  Heriot-Watt University\\
  Edinburgh, EH14 4AS, UK\\
  \texttt{is33@hw.ac.uk}\\
  \And
  Sungjin Lee \\
  Microsoft Research \\
  One Microsoft Way\\
  Redmond, WA 98052, USA\\
  \texttt{sule@microsoft.com} \\
}

\begin{document}

\maketitle

\begin{abstract}
Neural network-based dialog models often lack robustness to anomalous, out-of-domain (OOD) user input which leads to unexpected dialog behavior and thus considerably limits such models' usage in mission-critical production environments. The problem is especially relevant in the setting of dialog system bootstrapping with limited training data and no access to OOD examples. In this paper, we explore the problem of robustness of such systems to anomalous input and the associated to it trade-off in accuracies on seen and unseen data. We present a new dataset for studying the robustness of dialog systems to OOD input, which is bAbI Dialog Task 6 augmented with OOD content in a controlled way.
We then present turn dropout, a simple  yet efficient negative sampling-based technique for improving robustness of neural dialog models. We demonstrate its effectiveness applied to Hybrid Code Network-family models (HCNs) which reach state-of-the-art results on our OOD-augmented dataset as well as the original one. Specifically, an HCN trained with turn dropout achieves state-of-the-art performance of more than \textbf{75\%} per-utterance accuracy on the augmented dataset's OOD turns and \textbf{74\%} F1-score as an OOD detector. Furthermore, we introduce a Variational HCN enhanced with turn dropout which achieves more than \textbf{56.5\%} accuracy on the original bAbI Task 6 dataset, thus outperforming the initially reported HCN's result.
\end{abstract}

\section{Introduction}
% Sungjin
% Data-driven dialog systems have recently passed the stage of open-ended academic research and are currently being turned into consumer products on a massive scale. As such, data-driven dialog building techniques are adopted in platforms like \textit{Google Dialogflow}, \textit{Apple SiriKit}, \textit{Amazon Alexa Skills Kit}, and \textit{Microsoft Cognitive Services}. However, most of those platforms' data-driven functionality is limited to Natural Language Understanding: user intent detection, named entity recognition, and slot filling. 
Data-driven approaches for building dialog systems have recently passed the stage of open-ended academic research and are adopted in platforms like \textit{Google Dialogflow}, \textit{Apple SiriKit}, \textit{Amazon Alexa Skills Kit}, and \textit{Microsoft Cognitive Services}. However, most of those platforms' data-driven functionality is limited to Natural Language Understanding: user intent detection, named entity recognition, and slot filling.
A more unified approach to dialog system bootstrapping ~--- end-to-end dialog learning~--- is still only emerging as a commercial service, e.g. %as Microsoft Cognitive Services' prototype project
\textit{Microsoft Conversation Learner}. Although still in its early age, end-to-end dialog learning from examples offers great potential: it doesn't require advanced programming skills and thus it makes it possible for a wider range of users to create dialog systems for their purposes. In turn, in the enterprise environment, end-to-end dialog learning bridges the gap between user experience designers and the actual working systems thus making product cycles and overall workflow faster.

From the technical point of view, the key issue in end-to-end training is the lack of robustness of the resulting systems. 
% Unlike the traditional machine learning approach widely adopted in research projects, 
In the real-world setting of rapid dialog system prototyping, it is common to have only in-domain (IND) data for a closed target domain.
% there is a very limited amount of training data available. 
This leads to a significant overfitting of machine learning methods and unpredictable behavior in the cases outside of what was seen during training. For a closed-domain dialog system, it's extremely important to maintain predictable behavior on anomalous, OOD user input. 

In this paper, we focus on studying the effect of OOD input on end-to-end goal-oriented dialog models' performance and propose a simple and efficient solution to improving robustness only using IND data. Our contribution is thus two-fold: 
\begin{itemize}
    \item We present a dataset for studying the effect of OOD input on dialog models.
    \item We present turn dropout, an efficient negative sampling-technique for training dialog models that are capable of OOD handling using only the IND data.
\end{itemize}
% firstly, we present a dataset for studying the effect of OOD input on dialog models; secondly, we present turn dropout, an efficient negative sampling-technique for directly training for OOD handling in a supervised way while only using the IND data available. 
We show that HCN-based models enhanced with turn dropout show superior performance on OOD input, as well as surpass original HCN's result on IND-only data. 

\section{Related work}
Detection of anomalous input is a key research problem in machine learning. In the area of dialog systems, there is a series of approaches to detecting and processing of OOD input. If treated as a classification problem, this problem require both IND and OOD data~\citep{nakano2011two,tur2014detecting}. Although for the real-world scenario of end-to-end dialog system learning the task of collecting data covering potentially unbounded variety of OOD input is impractical. In contrast, there are also approaches like an in-domain verification method \citep{lane2007out}  and an autoencoder-based OOD detection \citep{ryu2017neural} which do not require OOD data. However, they still have restrictions such that there must be multiple sub-domains to learn utterance representation and one must set a decision threshold for OOD detection. For a dialog system that is supposed to work in a single closed domain, these methods are not a viable solution.

In contrast to those approaches, we present a simple and efficient technique for training dialog systems robust to OOD input in an end-to-end way, which allows the model to leverage the dialog context information to avoid the necessity of using IND data.
% , taking advantage of the dialog context information and integrated into the main system's training stage.

\section{Dataset for studying robustness of dialog systems}

\begin{figure}[t]
  \centering
  \includegraphics[width=0.9\linewidth]{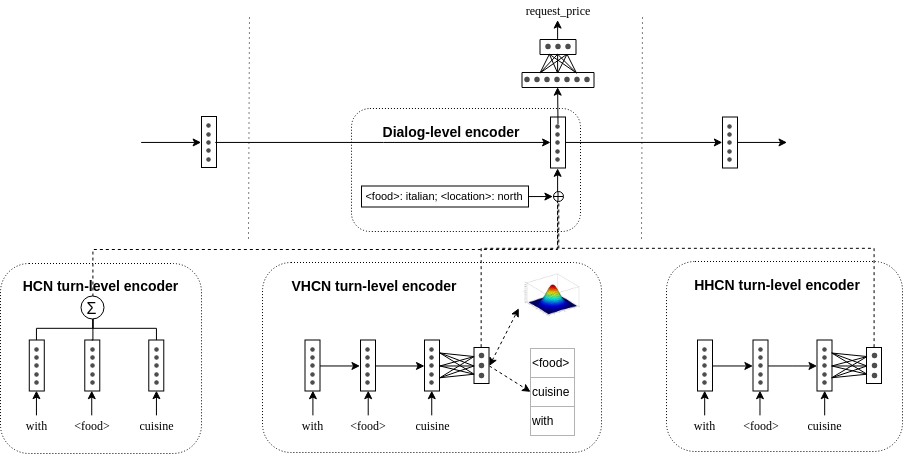}
  \caption{Hybrid Code Network model family}
  \label{fig:hcn_all}
\end{figure}

In order to study the effect of OOD input on end-to-end dialog system's performance, we used a dataset of real human-computer goal-oriented dialogs and augmented it with real user utterances from other domains in a controlled way using the open-source toolkit bAbI tools\footnote{\url{https://bit.ly/babi_tools}} \citep{Shalyminov.etal17}.

As our main dataset, we use bAbI Dialog Task 6 \citep{babi}, real human-computer conversations in the restaurant serach domain initially collected for Dialog State Tracking Challenge 2 \citep{DBLP:conf/sigdial/HendersonTW14}.

Our OOD augmentations are as follows:
\begin{itemize}
  \item \textit{turn-level OOD}: user requests from a foreign domain~--- the desired system behavior for such input is the fallback action,
  \item \textit{segment-level OOD}: interjections in the user in-domain requests~--- treated as valid user input and is supposed to be handled by the system in a regular way.
\end{itemize}

These two augmentation types reflect a specific dialog pattern of interest (see Table \ref{tab:augmentation_example}): first, the user utters a request from another domain at an arbitrary point in the dialog (each turn is augmented with the probability $p_{ood\_start}$), and the system answers accordingly. This may go on for several turns in a row~---each following turn is augmented with the probability $p_{ood\_cont}$. Eventually, the OOD sequence ends up and the dialog continues as usual, with a segment-level OOD of the user affirming their mistake. For this study, we set $p_{ood\_start}$ to $0.2$ and $p_{ood\_cont}$ to $0.4$\footnote{We experimented with other values of $p_{ood\_start}$ and $p_{ood\_cont}$ but didn't see significant differences in the results. Further experiments for different domains are encouraged using the tools provided}. 

While we introduce the OOD augmentations in a controlled programmatic way, the actual OOD content is natural. The turn-level OOD utterances are taken from dialog datasets in several foreign domains:
\begin{itemize}
    \item Frames dataset \citep{DBLP:conf/sigdial/AsriSSZHFMS17}~--- travel booking (1198 utterances),
    \item Stanford Key-Value Retrieval Network Dataset \citep{DBLP:conf/sigdial/EricKCM17}~--- calendar scheduling, weather information retrieval, city navigation (3030 utterances),
    \item Dialog State Tracking Challenge 1 \citep{DBLP:conf/sigdial/WilliamsRRB13}~--- bus information (968 utterances).
\end{itemize}

In order to avoid incomplete/elliptical phrases, we only took the first user's utterances from the dialogs.

For segment-level OOD, we mined utterances with the explicit affirmation of a mistake from Twitter and Reddit conversations datasets (e.g. ``my mistake'', ``I'm so sorry'')~--- 701 and 500 utterances respectively.
Our datasets, as well as the tools for OOD-augmentation of arbitrary datasets of interest are openly available\footnote{See \url{https://github.com/ishalyminov/ood_robust_hcn}}.

%The exact OOD pattern as described above is chosen in order to find out whether OOD input influences the states of dialog RNN-based encoders in a harmful way. More specifically, we test the following hypotheses:
%\begin{itemize}
%  \item how does turn-level OOD affect the system's performance further in the dialog,
%  \item how does it depend on the length of an OOD sequence,
%  \item how do turn and segment-level OOD contribute to that?
%\end{itemize}

\section{A data-efficient technique for training robust dialogue systems}

\begin{table*}[t]
  \centering
  \small
    \begin{tabular}{c|l|l}
      1&\textit{sys:}&Hello, welcome to the Cambridge restaurant system. You can ask for restaurants by area, \\
      &&price range or food type. How may I help you?  \\
      2&\textit{usr:}&\textbf{I'd like to get away from Monday, August 15, 2016 to Wednesday, August 31, 2016.} \\
      &&\textbf{My budget is \$3200 and I would leave from Detroit.} \\
      3&\textit{sys:}&Sorry I didn't catch that. Could you please repeat? \\
      4&\textit{usr:}&\textbf{Will there be frost in Durham next week?}\\
      3&\textit{sys:}&Sorry I didn't catch that. Could you please repeat? \\
      4&\textit{usr:}&\textbf{\textit{So sorry man}} I want a moderately priced restaurant in the west part of town \\
      5&\textit{sys:}&$<$api call$>$
    \end{tabular}
    \caption{Augmented dialog example (OOD content in bold, segment-level in italics)}
    \label{tab:augmentation_example}
\end{table*}

\subsection{Models}
In this paper, we experiment with Hybrid Code Network family of models \citep{DBLP:conf/acl/WilliamsAZ17}. HCN is reported to be state-of-the-art for the original, IND-only bAbI Dialog Task 6 data. Thus, in this paper we experiment with it and explore its robustness to OOD input.

HCN is a hierarchical dialog control model with a turn-level and a dialog-level components (we will call them both encoders). The turn-level encoder produces a latent representation of a single dialog turn, and the dialog-level one augments it with additional dialog-level features such as binary indicators of which slot values have been provided and whether the latest API call returned any results. Dialog-level encoder (RNN-based for all the models described) outputs a latent representation of the entire dialog which is then fed into the predictor MLP. Its output is the sequence of dialog actions (restricted by binary action masks provided by domain experts). Our models are described below~--- they share the same dialog-level encoder and predictor. The differences are on the turn level and in the overall optimization objective (see Figure \ref{fig:hcn_all} for an illustration).

\textbf{HCN}~--- the original model introduced by \citep{DBLP:conf/acl/WilliamsAZ17}. Its encoding of the user's input turn $x$ consisting of $N$ tokens is as follows:
\begin{equation} \label{eq:w2v}
HCN(x) = \frac{1}{N} \sum_i{w2v(x_i)}
\end{equation} where $w2v$ is the pre-trained Google News word2vec embeddings (frozen at the training time). HCN's optimization objective is categorical cross-entropy with respect to log-likelihood (here and in Eq. \ref{eq:l_vhcn} we show maximization objectives for simplicity. In the actual implementation, they are minimized with their sign reversed):
\begin{equation} \label{eq:l_hcn}
\mathcal{L}_{HCN} = \log p(a|x, c)
\end{equation} where $a$ is the dialog action and $c$ is dialog context.

\textbf{Hierarchical HCN (HHCN)} uses an RNN (in our case an LSTM cell \citep{lstm}) for encoding each utterance:
\begin{equation} \label{eq:hhcn}
HHCN(x) = LSTM(x)
\end{equation}
The optimization objective is the same as of HCN. Variants of this model were described by \citep{DBLP:journals/corr/abs-1712-09943} and \citep{10.1007/978-3-319-95933-7_24}.
 
\textbf{Variational HCN (VHCN)} which, to the best of our knowledge, is presented here for the first time~--- uses a Variational Autoencoder as the turn-level encoder, so that the resulting turn encoding is VAE's latent variable (normally referred to as $z$):
\begin{equation}
VHCN(x) = \mu(LSTM(x)) + \sigma(LSTM(x)) * N(0, 1)
\end{equation}
 
Where $\mu$ and $\sigma$ are MLPs for predicting $z$'s posterior distribution parameters, and $N(0, 1)$ is a sample from its prior distribution, a standard Gaussian \citep{DBLP:conf/conll/BowmanVVDJB16}.
 
This model differs from the previous two in that it learns dialog control and autoencoding jointly. In order to keep the secondary task less complex than the main one, we represent VAE's reconstruction targets as bags of words (BoW). Thus, VHCN optimization objective is as follows:
\begin{equation} \label{eq:l_vhcn}
 \mathcal{L}_{VHCN} = \mathbb{E}_{q(z)}[log(p(a \mid z, c))] + \mathbb{E}_{q(z)}[p(x_{BoW} \mid z)] - KL(q(z\mid x) \mid\mid p(z))
 \end{equation}
 
In the above formula, the first term is the main task's log-likelihood of the dialog action $a$, the second one is the VAE's reconstruction term for the user input in the bag-of-words form $x_{BoW}$, and the last turn is $KL$-divergence between the prior and posterior distribution of the VAE's latent variable $z$~--- following \citep{DBLP:conf/conll/BowmanVVDJB16}, we compute it in a closed form.

Another benefit of the BoW loss is, as reported in \citep{DBLP:conf/acl/ZhaoZE17}, it helps keep the variational properties of the model (i.e. non-zero KL-term) without the necessity of using the KL-term annealing trick \citep{DBLP:conf/conll/BowmanVVDJB16} which is itself challenging to control in practice. Unlike the authors of the original BoW loss approach, we don't stack softmax cross-entropy losses for each token and instead use a single sigmoid cross-entropy loss for the entire BoW vector.

All the models above use the same dialog-level LSTM encoder with additional features concatenated to the turn representations: BoW turn features, dialog context features, and previous system action\footnote{Without the loss of the architecture generality, we have action mask vectors as additional features for the dialog-level LSTM \citep{DBLP:conf/acl/WilliamsAZ17}, but they don't convey any information and are always set to 1's}.

\subsection{Turn dropout}
In order to train a system robust to OOD in the absence of real OOD examples, we employ a negative sampling-based approach and generate them synthetically from available IND data with a technique we call \textit{turn dropout}. Namely, we replace random dialog turns with synthetic ones, and assign them the fallback action.

More formally, our dialog features are as follows: $<f\_turn, f\_ctx, f\_mask, a>$, i.e. turn features (token sequences), dialog context features, action masks, and target actions respectively.

Under turn dropout, for a randomly selected dialog $i$ and its turn $j$, we replace $f\_turn_{ij}$ with a sequence of random vocabulary words (drawn from a uniform distribution over the vocabulary) and UNK tokens, and corresponding $a_{ij}$ with the fallback action, and leave all other features intact. In this way, we're simulating anomalous turns for the system given usual contexts (as stored in the dialog RNN's state), and we put minimum assumptions on the synthesized turns' structure (we only limit their lengths to be within the bounds of the real utterances).
\section{Experimental setup and evaluation}
\label{sec:evaluation}

\begin{table*}[t]
  \centering
  \small
    \begin{tabular}{|l||c||c|c|c|c|}
      \hline
      \textbf{Model}&\textbf{bAbI Dialog Task 6}&\multicolumn{4}{c|}{\textbf{bAbI Dialog Task 6 + OOD}}\\
      &Overall acc.&\multicolumn{1}{c}{Overall acc.}&\multicolumn{1}{c}{Seg. OOD acc.}&\multicolumn{1}{c}{OOD acc.}&OOD F1\\\hline\hline
      HCN&0.557&0.438&\textbf{0.455}&0.0&0.0\\\hline
      HHCN&0.531&0.418&0.424&0.0&0.0\\\hline
      VHCN&0.533&0.413&0.413&0.0&0.0\\\hline\hline
      TD-HCN&0.563&\textbf{0.575}&0.257&\textbf{0.754}&\textbf{0.743}\\\hline
      TD-HHCN&0.505&0.455&0.435&0.274&0.418\\\hline
      TD-VHCN&\textbf{0.565}&0.545&0.407&0.530&0.667\\\hline
    \end{tabular}
    \caption{Evaluation results}
    \label{tab:evaluation}
\end{table*}

We train our models only using the original bAbI Dialog Task 6 dataset, and evaluate them on our OOD-augmented versions of it: we use the per-utterance accuracy as our main evaluation metric; the models are trained with the same hyperparameters (where applicable) listed in Table \ref{tab:hcn_setup}. The models use the common unified vocabulary including all words from our datasets (including OOD content): the intuition behind this is as follows: production dialog models often use word embedding matrices with vocabularies significantly exceeding that of the training data in order to take advantage of additional generalization power via relations like synonymy, hyponymy, or hypernymy normally efficiently handled by distributed word representations. Therefore, mapping every unseen word to an `UNK' doesn't quite reflect that setting.

We tuned our models' hyperparameters using 2-stage grid search, tracking the development set accuracy. At the first stage, we adjusted the embedding dimensionality of our models (and the latent variable size in case of VHCN). Then, given the values found, at the second stage we adjusted turn dropout ratio at the interval $[0.05 - 0.7]$. Exact hyperparameter values are detailed in Table \ref{tab:hcn_setup}.  

The results are shown in Table \ref{tab:evaluation} ~--- please note, apart from the accuracies we report OOD F1-measure, a metric showing the model's performance as a conventional OOD detector, with positive class being the fallback action, and negative~--- all the IND classes actions.

Finally, given the stochastic nature of VHCN, we reported its mean accuracy scores over 3 runs (we used the same criterion for selecting the best model during the training procedure).

\section{Discussion and future work}
In this paper, we explored the problem of robustness of neural dialog systems to OOD input. Specifically, we presented a dataset for studying this problem along with a general procedure for augmenting arbitrary datasets of interest for such purpose. Secondly, we introduced turn dropout, a simple yet efficient technique for improving OOD robustness of dialog control models and evaluated its effect on several Hybrid Code Network-family models.

As our experiments showed, while learning to handle both IND and OOD input with access to IND-only data at the training time, there appears the following trade-off: a model performing better on the `clean' test turns is prone to lower accuracy on OOD~--- it can be said that it slightly overfits to its devset. On the other hand, a model regularized with turn dropout during training naturally performs better on unseen OOD turns, but with not as high accuracy on its `clean', IND test data. Another side of the trade-off is the accuracy of OOD detection vs robust handling of IND input with segment-level noise. As our results showed, models specifically trained for OOD detection all demonstrate lower accuracy on the noisy IND.

Among the models we evaluated, it's worth noting that the original HCN demonstrated the best performance as an OOD detector (more than \textbf{74\%} F1-score) and thus overall IND + OOD accuracy on the augmented dataset~--- more than \textbf{57\%}. While some parts of its architecture (e.g. mean vector-based turn encoding or bag-of-words feature vector at the utterance level) may not seem to be the most robust solution, the model demonstrate superior overall performance. Averaging at the turn level instead of recurrent encoding (the case of HHCN and VHCN) makes the model less dependent on actual word sequences seen during training but on the keywords themselves. 

In turn, VHCN demonstrated superior performance on IND data when trained with turn dropout, more than \textbf{56\%}~--- it benefited in terms of both overall accuracy and the absence of false-positive OODs thus outperforming the original HCN as reported by \citep{DBLP:conf/acl/WilliamsAZ17}. An additional challenge was to train it while keeping its variational properties (i.e. reasonably high KL term)~--- the BoW reconstruction loss which we used in order to simplify the secondary task, helped with this as well \citep{DBLP:conf/acl/ZhaoZE17}. On the other hand, while achieving superior performance on clean data, VHCN's properties didn't result in OOD handling improvements.

The question which is still unanswered is how these techniques apply to the setting of few-shot training. In the practical setup of training dialog systems from minimal data, having access to even medium-sized datasets like bAbI Dialog Task 6 isn't realistic, and all the initial requirements for the models have to be met only using the minimal training data available. It's the next step in our research to explore how our techniques apply to this setup and what needs to be done in order to achieve OOD robustness with maximum few-shot data efficiency.

\bibliographystyle{acl_natbib}
\bibliography{acl2018}

\begin{appendices}
\section{}
\begin{table*}[h!]
  \centering
  \small
    \begin{tabular}{|l|c|c|c|}
      \hline
      \textbf{Hyperparameter}&\textbf{HCN}&\textbf{HHCN}&\textbf{VHCN}\\\hline\hline
      Embedding size&64&128&128\\\hline
      Latent variable size&---&---&8\\\hline
      Learning rate&\multicolumn{3}{c|}{0.001}\\\hline
      Optimizer&\multicolumn{3}{c|}{Adam}\\\hline
      Early stopping threshold (epochs)&\multicolumn{3}{c|}{20}\\\hline
      Turn dropout ratio&0.4&0.6&0.3\\\hline
      Word dropout ratio&\multicolumn{3}{c|}{0.2}\\\hline
    \end{tabular}
    \caption{Model hyperparameters}
    \label{tab:hcn_setup}
\end{table*}

\end{appendices}
\end{document}